\documentclass[letterpaper, 10 pt, conference]{ieeeconf}  

\IEEEoverridecommandlockouts                              

\usepackage{cite}
\usepackage{amsmath,amssymb,amsfonts}
\usepackage{algorithmic}
\usepackage{graphicx}
\usepackage{textcomp}
\usepackage{xcolor}
\usepackage{svg}
\usepackage{layouts}
\usepackage{booktabs}
\usepackage{float}

\newcommand{\mb}[1]{\mathbf{#1}}

\title{\LARGE \bf
End-to-end Reinforcement Learning \\for Time-Optimal Quadcopter Flight
}

\author{Robin Ferede$^{1}$, Christophe De Wagter$^{1}$, Dario Izzo$^{2}$,  Guido C.H.E. de Croon$^{1}$
\thanks{*This work was supported by ESA}
\thanks{$^{1}$The authors are with the Micro Air Vehicle Lab of the Faculty
of Aerospace Engineering, Delft University of Technology, 2629 HS
Delft, The Netherlands {\tt\small R.Ferede@tudelft.nl, G.C.H.E.deCroon@tudelft.nl, C.deWagter@tudelft.nl}}%
\thanks{$^{2}$The authors are with the Advanced Concepts Team, European Space Agency, Keplerlaan 1, 2201 AZ, Noordwijk, The Netherlands. {\tt\small Dario.Izzo@esa.int}}%
}

\begin{document}

\maketitle
\thispagestyle{empty}
\pagestyle{empty}

\begin{figure*}
    \centering
    \includegraphics[width=\linewidth]{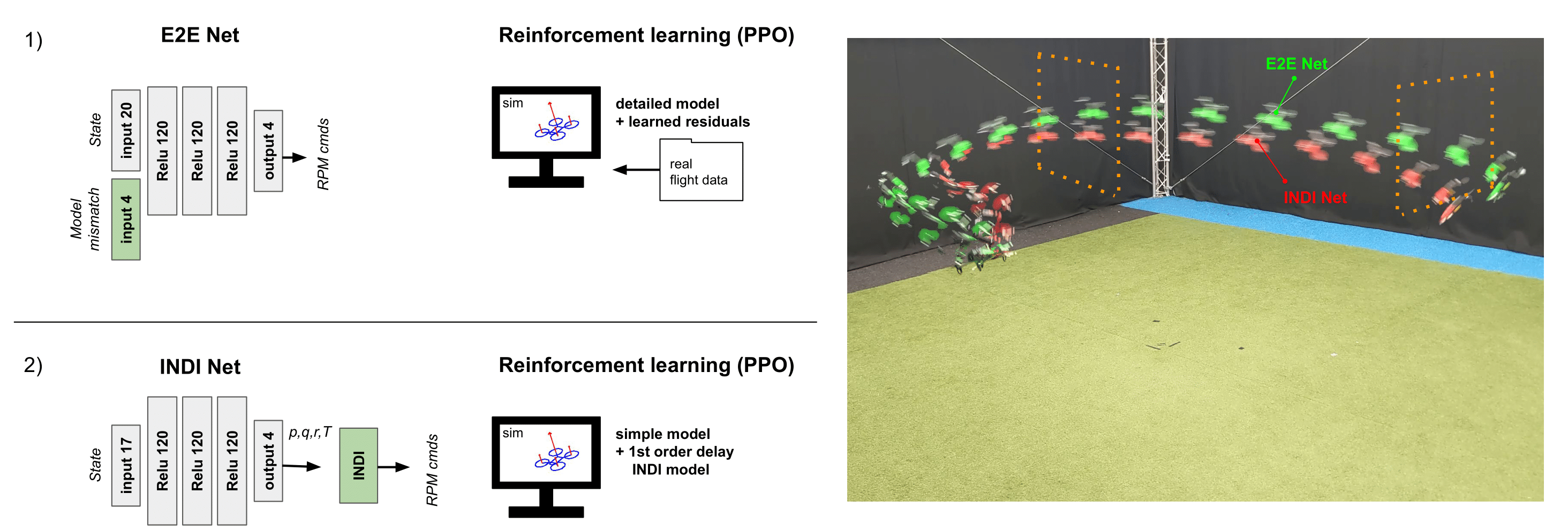}
    \caption{Drone racing with 2 RL Control Approaches: 1) E2E Network: Computes direct motor commands, learns compensation for unmodeled moments and thrust, trained with a learned residual model. 2) INDI Network: Computes thrust and body rate commands, uses an INDI inner loop controller, trained with first order INDI model.}
    \label{fig:Fig1}
\end{figure*}
\begin{abstract}
Aggressive time-optimal control of quadcopters poses a significant challenge in the field of robotics. The state-of-the-art approach leverages reinforcement learning (RL) to train optimal neural policies. However, a critical hurdle is the sim-to-real gap, often addressed by employing a robust inner loop controller —an abstraction that, in theory, constrains the optimality of the trained controller, necessitating margins to counter potential disturbances. In contrast, our novel approach introduces high-speed quadcopter control using end-to-end RL (E2E) that gives direct motor commands. To bridge the reality gap, we incorporate a learned residual model and an adaptive method that can compensate for modeling errors in thrust and moments. We compare our E2E approach against a state-of-the-art network that commands thrust and body rates to an INDI inner loop controller, both in simulated and real-world flight. E2E showcases a significant 1.39-second advantage in simulation and a 0.17-second edge in real-world testing, highlighting end-to-end reinforcement learning's potential. The performance drop observed from simulation to reality shows potential for further improvement, including refining strategies to address the reality gap or exploring offline reinforcement learning with real flight data.

%

\end{abstract}

\section{INTRODUCTION}
The demand for autonomous quadcopters capable of high-speed flight has been steadily increasing, driven by the need for covering larger distances in various applications\cite{Hassanalian2017ClassificationsAA}. However, achieving efficient and agile high-speed flight remains a significant challenge, requiring the development of computationally efficient time-optimal control algorithms.

Traditionally, quadcopter control relied on established methods, including Differential-Flatness-Based Controllers (DFBC) \cite{NIEUWSTADT19962301, MinSnap, Faessler2018, tal2020accurate} and Nonlinear Model Predictive Control (NMPC) \cite{aerospace4020031, Bicego2020, explicitMPC, torrente2021data, MPCC, TimeOpimalReplanning, MPC_DFBP, AdaptiveNMPC, torrente2021data}. While these methods represented important steps forward, they faced difficulties with unmodeled effects \cite{OCvsRL}.

Recent advancements in quadcopter control have brought Reinforcement Learning (RL) to the forefront of research \cite{droneracesurvey}. While simulation studies such as \cite{nagami2021hjb, ates2020long} have demonstrated RL's capability to achieve high-speed flight, real-world applications such as \cite{Autonomous_Drone_Racing_with_Deep_Reinforcement_Learning, penicka2022learning, kaufmann2022benchmark, Kaufmann2023} have had to devise clever solutions to address the issues posed by the reality gap. This is because RL heavily relies on accurate simulation environments, which become increasingly challenging at high quadcopter speeds due to complex dynamics. Consequently, RL methods in actual flight often adopt an abstraction approach in which the network outputs high-level control commands that are executed by manually pre-tuned low-level controllers.

For instance, in \cite{Autonomous_Drone_Racing_with_Deep_Reinforcement_Learning}, an end-to-end network is trained in simulation for racing and then used in real flight to generate a trajectory tracked by an MPC controller. In \cite{penicka2022learning}, RL provides thrust and body rate commands, which are tracked with an inner loop controller. In a comparison study \cite{kaufmann2022benchmark}, it is argued that providing thrust and body rate commands strikes the optimal balance, as direct motor commands, while effective in simulation, fail to bridge the reality gap for successful real-life flight. Particularly noteworthy is the achievement in \cite{Kaufmann2023}, where an RL controller outperformed a human racing pilot. This success was attributed to a combination of abstraction (thrust and body rate commands) and accurate modeling of the closed-loop system using a learned residual model.

However, leaning on abstractions, in theory, places a limitation on the platform's maneuverability. This is because, in the end, it is the lowest-level controller that decides which actuator to saturate. On the contrary, methods like end-to-end (E2E) approaches, which operate without such abstractions, hold promise in enabling exceptionally agile maneuvers and pushing performance to its absolute limits. 

There are already some noteworthy instances of neural networks directly governing motor control. For instance, in \cite{hwangbo2017control}, direct motor control is achieved by a neural network, focusing on low-level controls in waypoints tracking and
vehicle recovery from harsh initialization. Although this work was concerned with end-to-end control, it did not focus on time optimality or high-speed flight. In more recent developments, methods using supervised learning from optimal trajectory datasets have shown remarkable success in E2E control in the context of high-speed quadcopter flight \cite{Ferede, Seb}. The strategies used here effectively address the reality gap by using an adaptive control approach that both measures and compensates for modeling errors.

In this article, we introduce a novel method for achieving high-speed quadcopter control through end-to-end reinforcement learning (E2E), where direct motor commands are generated. To effectively bridge the reality gap, our approach combines a residual moment and thrust model, acquired from flight data (akin to the approach in \cite{Kaufmann2023}), combined with the adaptive techniques described in \cite{Ferede,Seb}, applied to both thrust and moment model. Our study involves a comparative analysis, pitting the performance of our E2E approach against that of a state-of-the-art network generating thrust and rate commands for an inner loop INDI controller \cite{Smeur2016}.

\section{METHODOLOGY}
\subsection{Problem Definition}
The drone racing problem is modeled as an MDP defined by \((\mathcal{S}, \mathcal{A}, \mathcal{P}, r, \rho_0, \gamma)\). The RL agent starts in state \(s_t \in \mathcal{S}\) from distribution \(\rho_0\). It uses stochastic policy \(\pi(a_t|s_t)\) to select a continuous action \(a_t \in \mathcal{A}\), transitioning to \(s_{t+1}\) with probability \(\mathcal{P}_{s_t, s_{t+1}}^{a_t}\) and receiving reward \(r_t\). The objective is to optimize the parameters \(\theta\) of a stochastic neural network policy \(\pi_\theta(a_t|s_t)\) to maximize expected return over an infinite horizon \(\max_\theta \mathbb{E}_{\pi_\theta} \left[ \sum_{t=0}^{\infty} \gamma^t r_t \right]\).

The racing setup consists of four square gates (1x1m) in a 4x3m rectangle. Each gate is rotated 90 degrees relative to the previous gate as seen in Fig.~\ref{fig:top_view_comparison}. The goal is to fly in a circular trajectory, passing through all gates as quickly as possible. We quantify this by using a reward function consisting of a progress reward that quantifies advancement toward the target gate, a gate reward that encourages gate passage through the center, and a collision penalty.
\begin{align*}
    r(k) = \begin{cases}
        +10-10||\mb{p}_k - \mb{p}_{g_k}||, &\text{if gate passed} \\
        -10, &\text{if collision} \\
       ||\mb{p}_k - \mb{p}_{g_k}||-||\mb{p}_{k-1} - \mb{p}_{g_k}|| &\text{otherwise} \\
    \end{cases}
\end{align*}
Here, \(\mb{p}_{g_k}\) represents the position of the center of the current target gate, and \(\mb{p}_{k}\), \(\mb{p}_{k-1}\) are the drone's current and previous positions. A gate is deemed successfully passed when the drone intersects the 1 by 1 meter gate boundary. A collision is registered when the drone either makes contact with the ground or passes through the gate plane outside the gate boundary.
This reward function, in conjunction with a state transition model to be detailed in the subsequent section, will be employed to create a virtual environment and conduct training using the PPO algorithm \cite{ppo} in the Python library Stable-Baselines3 \cite{stable-baselines3}. During training, we will simulate 100 drones in parallel, using a discount factor $\gamma=0.999$ and a maximum episode duration of 12 seconds.

\begin{figure*}
    \centering
    \includegraphics[width=\linewidth]{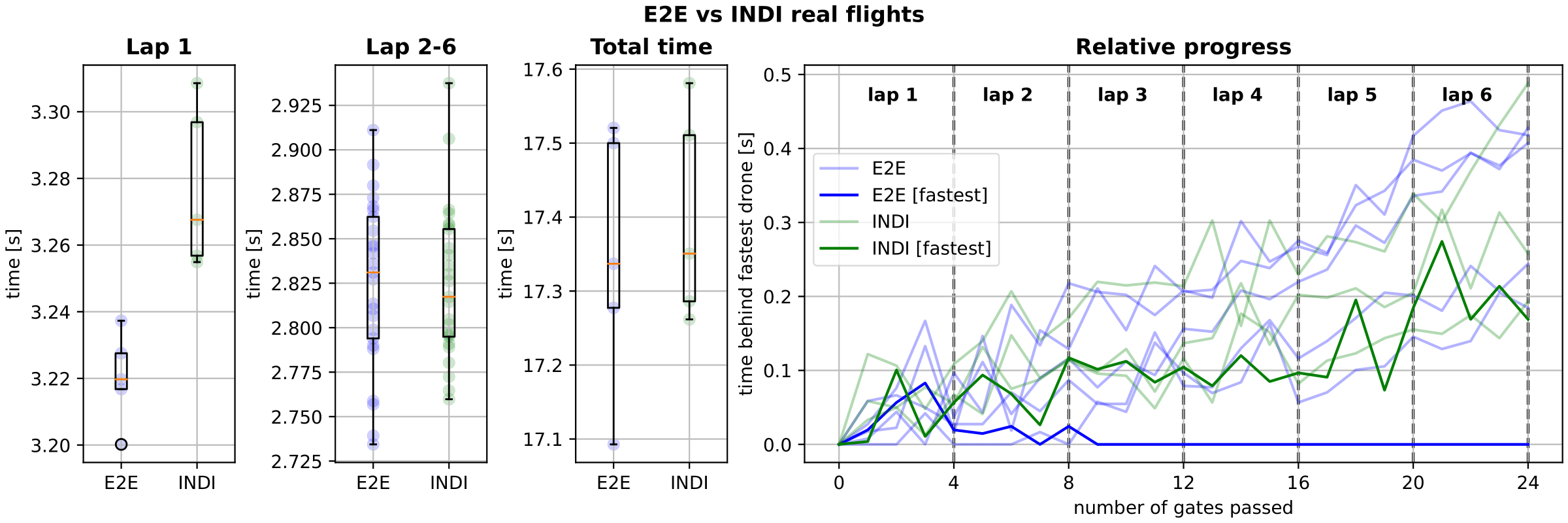}
    \caption{Flight test comparison E2E vs. INDI Net (5 Repetitions): 1) E2E outpaces INDI in the first lap (starting in hover). 2) Laps 2-6 show comparable performance. 3) E2E has a slight advantage over INDI in total track completion time. 4) Relative progress of all flights E2E leads by 0.17 seconds.}
    \label{fig:performance_comparison}
\end{figure*}

\subsection{E2E Network}
\subsubsection{Quadcopter model}
We adopt a quadcopter model similar to the one outlined in \cite{Ferede, Seb}. The quadcopter's state and control inputs are defined as follows:
\begin{align*}
    \mb{x} = [\mb{p}, \mb{v}, \boldsymbol \lambda, \mb{\Omega}, \mb{\boldsymbol \omega}, \mb{M}_{ext}, \mb{F}_{ext}]^T \quad \mb{u} = [u_1, u_2, u_3, u_4]^T
\end{align*}
Here $\mb{p}$ is the position of the drone, $\mb{v}$ is the velocity, $\boldsymbol \lambda$ are the Euler angles, $\bold \Omega$ are the body rates and $\boldsymbol \omega$ are the propeller speed in RPM with a range between $\omega_{min}=3000$ and $\omega_{max}=11000$. $\mb{u}$ represents the motor RPM commands. Similar to the approach in \cite{Ferede, Seb}, we include external (specific) forces and moments as part of the state representation. These external influences will be measured onboard to mitigate modeling inaccuracies. The equations of motion are expressed as follows:
\begin{align}
    \dot{\mb{p}} &= \mb{v} &
    \dot{\mb{v}} &= g \mb{e_3} + R(\mb{\lambda}) [\mb{F}_{mod} + \mb{F}_{ext}]\\
    \dot{\lambda} &= Q(\mb{\lambda}) \mb{\Omega} &
    I \dot{\mb{\Omega}} &= - \mb{\Omega} \times I \mb{\Omega} + \mb{M}_{mod}+\mb{M}_{ext} \nonumber \\
    \dot{\boldsymbol \omega}  &= (\mb{u} - \boldsymbol \omega)/\tau  &
    \dot{\mb{F}}_{ext} &= 0 \quad \dot{\mb{M}}_{ext} = 0\nonumber
\end{align}
In contrast to \cite{Ferede, Seb}, $\mb{M}_{mod}$ and $\mb{F}_{mod}$ are formulated as a hybrid model, comprising a nominal model rooted in physical principles \cite{bangura2016aerodynamics} and a residual black-box model fine-tuned using empirical data:
\begin{align*}
    \mb{F}_{mod} = \mb{F}_{nom} + \mb{F}_{res} \quad
    \mb{M}_{mod} = \mb{M}_{nom} + \mb{M}_{res}     
\end{align*}
The nominal model, identical to the one identified for the Bebop drone in \cite{Ferede}, is defined as:
\begin{align*}
    \mb{F}_{nom} = \begin{bmatrix}
        - k_x v^B_x \sum_{i=1}^4 \omega_i \\
        - k_y v^B_y \sum_{i=1}^4 \omega_i \\
        -k_\omega \sum_{i=1}^4 \omega_i^2 - k_z v^B_z \sum_{i=1}^4 \omega_i - k_h (v^{B2}_x + v^{B2}_y)
    \end{bmatrix}
\end{align*}
\begin{align*}
    \mb{M}_{nom} = \begin{bmatrix}
         k_p (\omega_1^2 - \omega_2^2 - \omega_3^2 + \omega_4^2) + k_{pv} v^{B}_y\\
         k_q (\omega_1^2 + \omega_2^2 - \omega_3^2 - \omega_4^2) + k_{qv} v^{B}_x\\
         k_{r1} (-\omega_1 + \omega_2 - \omega_3 + \omega_4) + \\
         \ldots k_{r2} (-\dot{\omega_1} + \dot{\omega_2} - \dot{\omega_3} + \dot{\omega_4})  - k_{rr} r
    \end{bmatrix}
\end{align*}
The residual model consists of two small neural networks, one for residual (mass normalized) thrust and the other for residual moment:
\begin{align*}
\mb{F}_{res} = [0, 0, -\mathcal{N}_{T}(\boldsymbol \omega, \mb{v}^B)]^T \quad
\mb{M}_{res} = \mathcal{N}_{M}(\boldsymbol \omega, \mb{v}^B, \boldsymbol \Omega)
\end{align*}
Both $\mathcal{N}_{T}$ (7 inputs 1 output) and $\mathcal{N}_{M}$ (10 inputs 3 outputs) utilize a neural network architecture featuring a single hidden layer comprising 32 neurons.

\subsubsection{Policy}
The policy is implemented as a three-layered, fully connected neural network with a ReLU activation function, as visually depicted in Fig.~\ref{fig:Fig1}. The network takes 24 inputs containing the quadcopter's state, as well as the modeling mismatches represented by \(\mb{M}_{ext}\) and \(\mb{F}_{ext}\), along with information current and future gates.
\begin{align*}
    \mb{x}_{in} = [\mb{p}^{g_i}
, \mb{v}^{g_i}, \boldsymbol \lambda^{g_i}, \mb{\Omega}, \mb{\boldsymbol \omega}, \mb{M}_{ext}, \mb{F}_{ext}, \mb{p}_{g_{i+1}}^{g_{i}}, \psi_{g_{i+1}}^{g_i}]^T
\end{align*}
Here, the superscript $^{g_i}$ denotes that the variables are represented in the reference frame aligned with the target gate $g_i$. The symbols $\mb{p}_{g{i+1}}$ and $\psi_{g_{i+1}}$ respectively indicate the position and orientation of the succeeding gate (i.e., the one following the target gate). The network has 4 outputs corresponding to rpm commands $\mb{u}$. Since our approach employs a stochastic policy, the network's outputs serve as the means of a normal distribution governing the rpm commands. The standard deviation of this distribution is a learned parameter of the network. For the purpose of generating a control command, whether during training or testing, we draw samples from this distribution and constrain the result within the minimum and maximum rpm limits.
\begin{figure*}
    \centering
    \includegraphics[width=\textwidth]{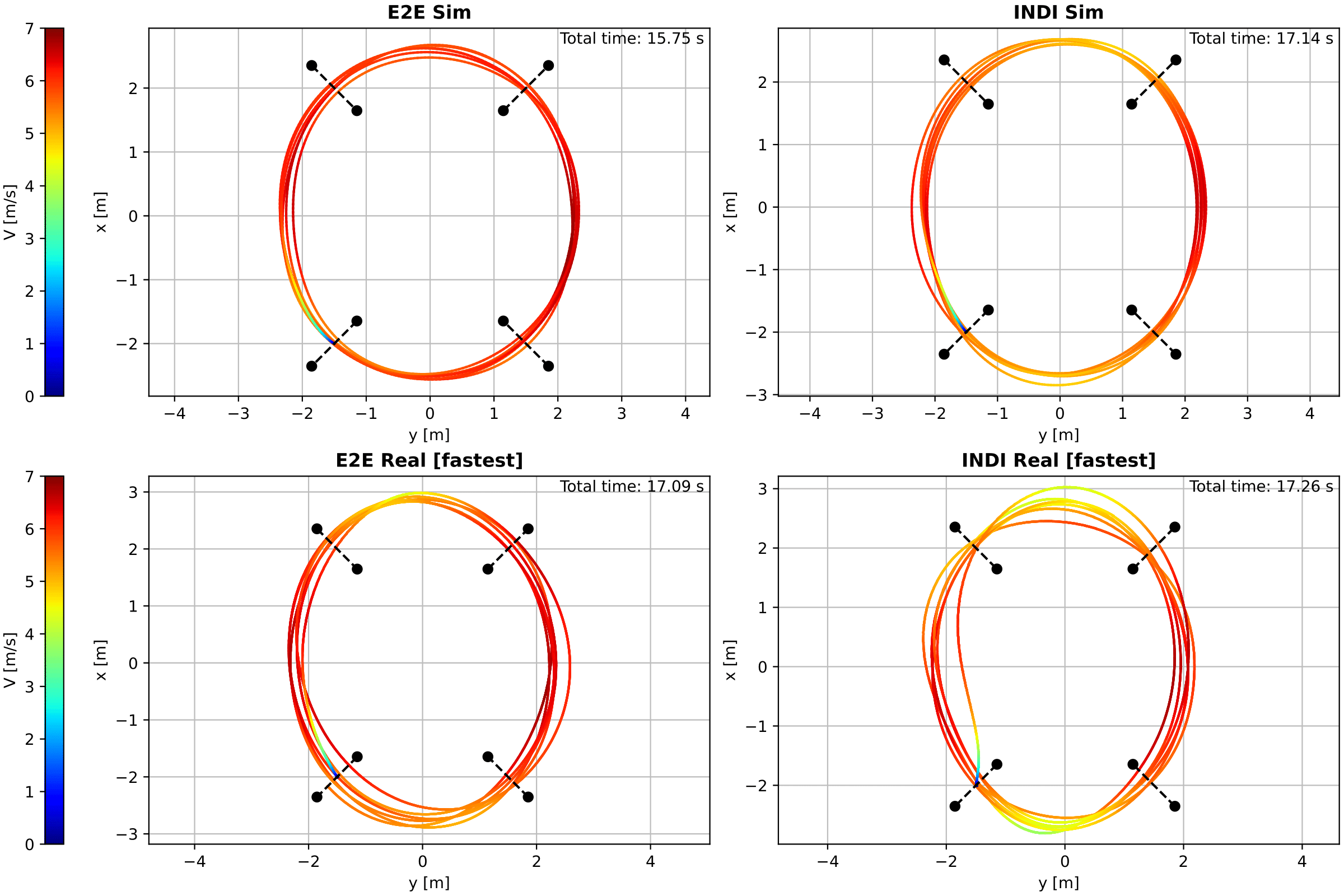}
    \caption{Trajectory comparison E2E vs INDI: simulated runs and fastest achieved real flights. The performance gap is most significant in simulation with E2E leading by 1.39 seconds. In the real flights, this difference is reduced to 0.17 seconds.}
    \label{fig:top_view_comparison}
\end{figure*}
\subsubsection{Training}
To train within the RL framework, we convert our continuous-time dynamical model into a discrete-time MDP using a time step of 0.01 via the Forward Euler method. The episode is initialized by uniformly sampling initial states from the following intervals:
\begin{align} \label{eq:initial_states}
    x &\in [-\tfrac{1}{2},\tfrac{1}{2}]+x_s &
    y &\in [-\tfrac{1}{2},\tfrac{1}{2}]+y_s &
    z &\in [-\tfrac{1}{2},\tfrac{1}{2}]+z_s \nonumber \\
    v_x &\in [-\tfrac{1}{2},\tfrac{1}{2}] &
    v_y &\in [-\tfrac{1}{2},\tfrac{1}{2}] &
    v_z &\in [-\tfrac{1}{2},\tfrac{1}{2}] \nonumber \\
    \phi &\in [-\tfrac{2 \pi}{9},\tfrac{2 \pi}{9}] &
    \theta &\in [-\tfrac{2 \pi}{9},\tfrac{2 \pi}{9}] &
    \psi &\in [-\pi, \pi] \nonumber \\
    p &\in [-1,1] &
    q &\in [-1,1] &
    r &\in [-1,1] \nonumber \\
    \boldsymbol \omega &\in [\omega_{min},\omega_{max}]^4 
\end{align}
Here $x_s,y_s,z_s$ is the starting point of the race track. Also \(\mb{M}_{ext}\) and \(\mb{F}_{ext}\) are sampled from
\begin{align*}
    M_{x,ext},M_{y,ext} &\in [-0.03, 0.03] &
    M_{z,ext} &\in [-0.01, 0.01] \\
    F_{x,ext},F_{y,ext} &= 0 &
    F_{z,ext} &\in [-0.5, 0.5]
\end{align*}

\subsection{INDI Network}
\subsubsection{Quadcopter model}
Integrating the INDI controller into the loop streamlines the quadcopter modeling process considerably. With the INDI Network providing rate and thrust commands, our RL environment now primarily focuses on simulating the INDI controller's response, specifically how actual thrust and rates relate to the commanded values. As illustrated in \cite{Smeur2016}, the INDI controller's command response can be expressed as a first-order delay model. The remaining quadcopter modeling tasks only involve the kinematic equations and a basic drag model:
\begin{align*}
    \mb{x} = [\mb{p}, \mb{v}, \boldsymbol \lambda, \mb{\Omega}, T]^T \quad \mb{u} = [\mb{\Omega}_{cmd}, T_{cmd}]^T
\end{align*}
The equations of motion are expressed as:
\begin{align}
    \dot{\mb{p}} &= \mb{v} &
    \dot{\mb{v}} &= g \mb{e_3} + R(\mb{\lambda}) \mb{F}\\
    \dot{\lambda} &= Q(\mb{\lambda}) \mb{\Omega} &
    \dot{\mb{\Omega}} &= (\mb{\Omega}_{cmd} - \mb{\Omega})/\tau_{\Omega} \nonumber \\
    \dot{T}  &= (T_{cmd}-T)/\tau_T  & & \nonumber
\end{align}
Here, we have identified the thrust and rate delays of the INDI controller, $\tau_{\Omega}$ and $\tau_T$, as 0.03 based on flight data. The specific force $\mb{F}$ is given by $\mb{F} = [-d_x v_x^B, -d_y v_y^B, -T]^T$ where $d_x=0.34$ and $d_y=0.43$ are also identified from flight data. 

An inherent limitation of relying on an INDI inner loop controller is that it requires some margins to reject disturbances and track the desired command. This means that a trade-off must be made between how much margin is kept to reject disturbances (which makes the drone slower) or how little margin is acceptable before actuator limits are exceeded and the INDI can not track the reference anymore. We establish artificial boundaries for thrust and rate based on the fastest stable flight performance we could achieve in our setup:
\begin{align*}
    p_{cmd}, q_{cmd}, r_{cmd} \in [-3,3] \quad T_{cmd} \in [0,15]
\end{align*}
\subsubsection{Policy}
The policy closely resembles the E2E Network but differs only in the number of inputs due to the system simplifications. The network takes 17 inputs, consisting of the state and gate information:
\begin{align*}
    \mb{x}_{in} = [\mb{p}^{g_i}
, \mb{v}^{g_i}, \boldsymbol \lambda^{g_i}, \mb{\Omega}, T, \mb{p}_{g_{i+1}}^{g_{i}}, \psi_{g_{i+1}}^{g_i}]^T
\end{align*}
The network has 4 outputs corresponding to the thrust and body rate control command. Again the network's outputs are the means of normal distributions governing thrust and body rate commands, along with a learned standard deviation.
\subsubsection{Training}
In the training phase, we once more convert our dynamical model into a discrete-time MDP. We uniformly sample initial states from the same intervals as described in Eq.~\ref{eq:initial_states}, where applicable. The thrust is sampled from the range [7.4, 7.6].
\section{EXPERIMENTAL SETUP}
For our experiments, we use the Parrot Bebop 1 quadcopter. While it does not have the high-speed performance of a racing drone, its unique characteristics make it a compelling choice. The Bebop 1 features a notably flexible frame, posing a more intricate challenge for modeling compared to the rigid racing counterparts. Furthermore, it provides an excellent opportunity to push the boundaries of actuation within a confined space. The onboard software of the Bebop is replaced by the Paparazzi-UAV open-source autopilot \cite{Gati2013}. Computation occurred in real-time on a Parrot P7 dual-core CPU Cortex A9 processor. The Bebop featured an MPU6050 IMU sensor for specific force and angular velocity measurements, as well as RPM measurements for each propeller, essential for our control method.
\begin{figure}
    \centering    \includegraphics[width=\linewidth]{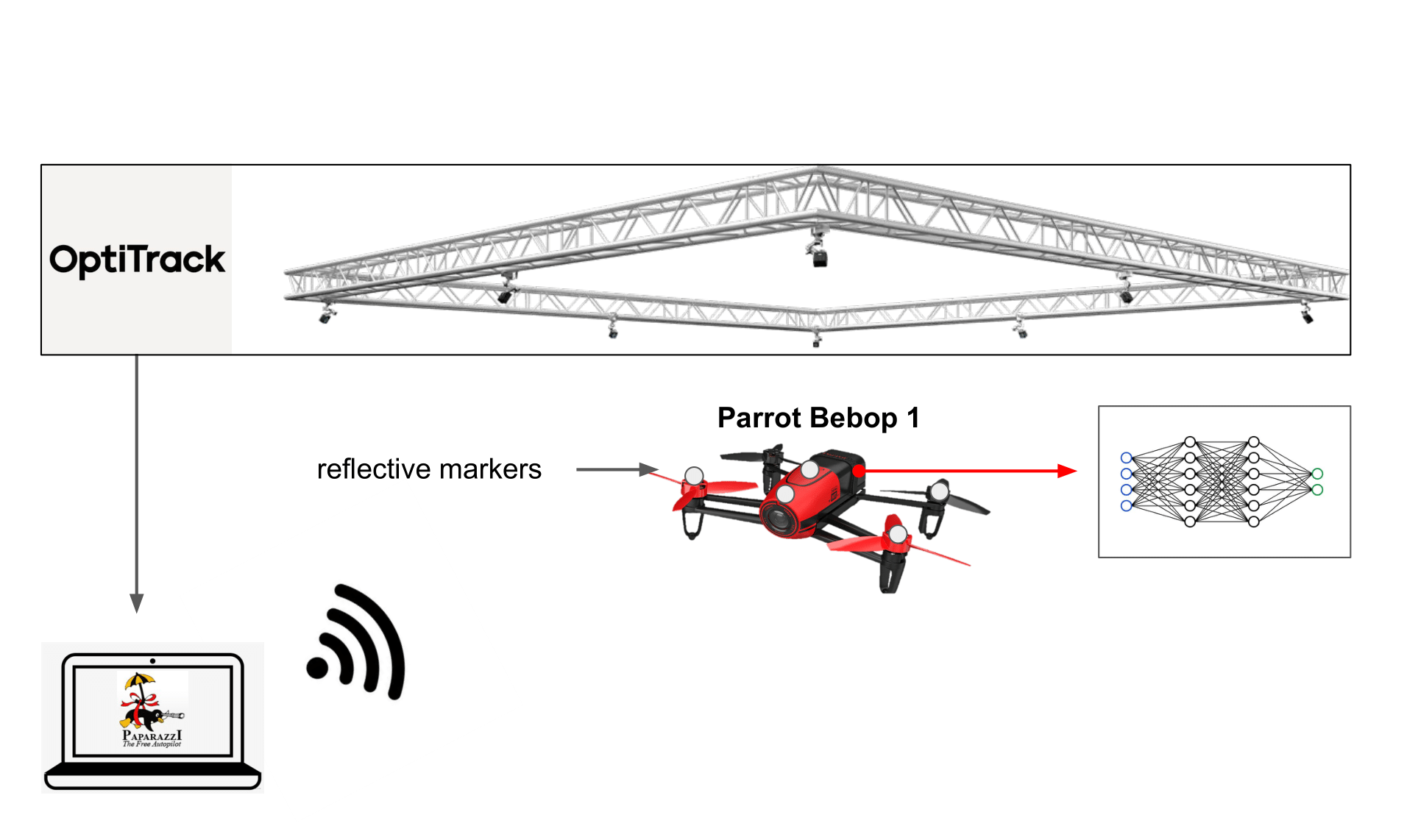}
    \caption{Experimental setup: The drone's position and attitude are obtained from the Optitrack motion capture system and fused with IMU data by an extended Kalman filter. Image from \cite{Ferede}}
    \label{fig:experimental_setup}
\end{figure}
Flight tests were conducted in The CyberZoo, a 10x10x7 m flight arena at TU Delft's Aerospace Engineering faculty, equipped with an OptiTrack motion capture system for real-time position and attitude data. An extended Kalman filter is used to fuse Optitrack position and attitude measurements with the accelerometer and gyro data from the IMU. This integration allows us to estimate position, velocity, attitude, and IMU biases accurately.

\section{RESULTS \& DISCUSSION}
We conducted tests using the trained E2E and INDI networks, involving both simulation roll-outs and real flight tests covering 6 laps on the race track. The real flights were repeated 5 times for both networks. Our assessment focused on time optimality as the primary performance metric. Fig.~\ref{fig:performance_comparison} provides an overview of the real flights conducted, along with corresponding values found in Table~\ref{tab:lap_times}. Fig.~\ref{fig:top_view_comparison} highlights the fastest flights achieved by both controllers compared to their simulated counterparts.

Our key observation is that E2E outperforms INDI on the 6-lap racing task, both in simulation and in real-world scenarios. The most substantial performance gap is evident in simulation, with E2E completing the track 1.39 seconds faster. In the real world, the performance difference, while reduced, remains existent, with an average improvement of 0.05 seconds over the entire track. When comparing the fastest runs, E2E is in the lead by 0.17 seconds. An intriguing insight, highlighted in Fig.~\ref{fig:performance_comparison}, reveals that E2E Networks primarily leverages its advantage during the first lap when the drone starts in hover. In laps 2 through 6, both methods exhibit remarkably similar performance. This observation highlights the E2E Net's capability to efficiently perform rapid accelerations.

When comparing the results between simulated runs and real-life flights, several noteworthy observations come to light. In the simulated environment, the INDI controller achieves a track completion time of 17.14, with only a minor increase to 17.40 seconds on average in the real flight tests. In contrast, the E2E method exhibits a more pronounced reality gap, with completion times increasing from 15.75 in simulation to an average of 17.35 in real-world scenarios. Despite this substantial gap between simulation and reality, the fact that E2E maintains a performance advantage over INDI, albeit modest, underscores the success of the sim-to-real transfer. Furthermore, these findings suggest that E2E may hold untapped potential for further enhancing real-world drone performance.

\begin{figure}
    \centering
    \includegraphics[width=\linewidth]{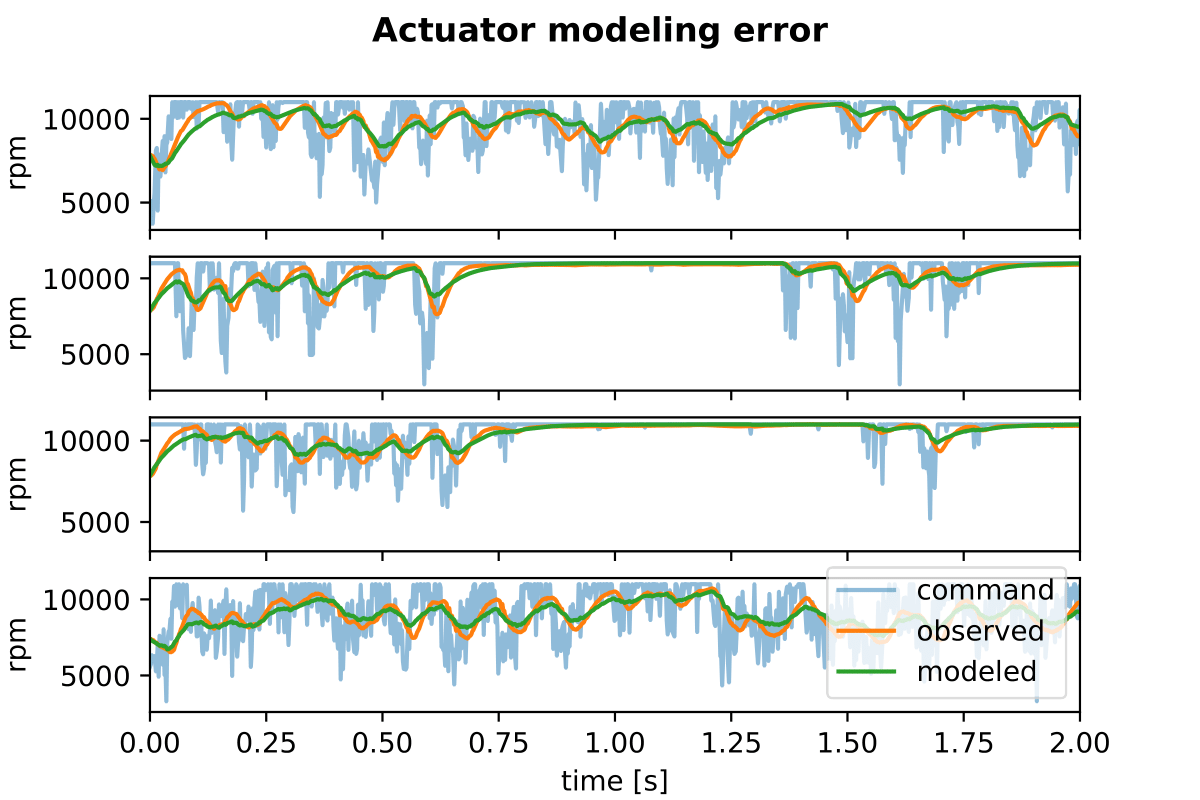}
    \caption{Modeling errors in the first order actuator delay model used for the E2E Net}
    \label{fig:actuator_errors}
\end{figure}

\begin{figure}
    \centering
    \includegraphics[width=\linewidth]{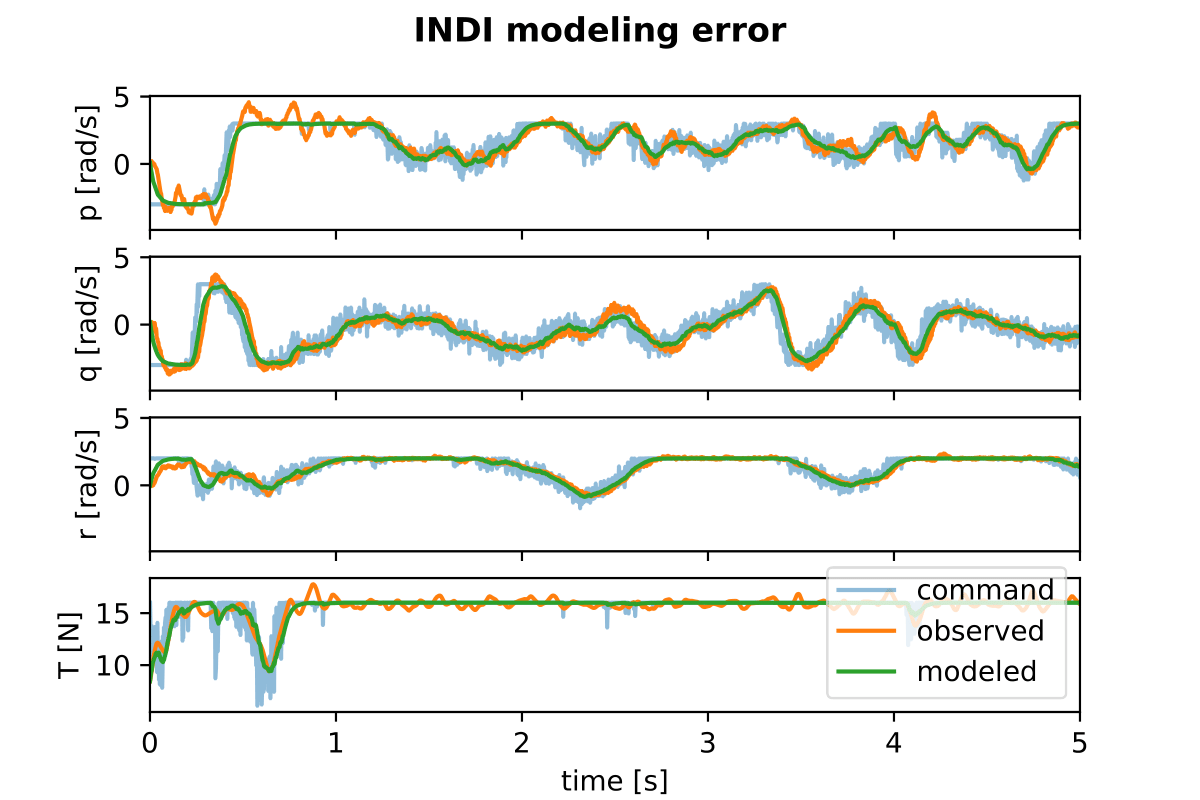}
    \caption{Modeling errors in the first order delay model of the INDI controller used for the INDI Net.}
    \label{fig:indi_errors}
\end{figure}

While the E2E net addresses thrust and moment modeling errors, it's important to note that certain challenges remain unaddressed. In Fig.~\ref{fig:actuator_errors}, we observe deviations in motor RPM between actual and modeled values during the fastest flight, potentially explaining performance drops in the real world. The INDI Net has a smaller simulation-to-reality gap, but it also encounters modeling errors. In Fig.~\ref{fig:indi_errors} we can see how the observed rates and thrust also deviate from the modeled first-order delay. These deviations could explain the different trajectory shapes observed in the real INDI flight in Fig.~\ref{fig:top_view_comparison}.

\begin{table}
    \centering
\begin{tabular}{|c|c|c|c|c|c|c|c|c|c|c|c|}
\toprule
 & lap1 & lap2 & lap3 & lap4 & lap5 & lap6 & total \\
\midrule
\hline
E2E Sim & 2.97 & 2.51 & 2.56 & 2.55 & 2.59 & 2.57 & 15.75 \\
\hline
INDI Sim & 3.20 & 2.82 & 2.75 & 2.80 & 2.81 & 2.76 & 17.14 \\
\hline
\hline
E2E1 & 3.23 & 2.79 & 2.81 & 2.84 & 2.84 & 2.83 & 17.34 \\
\textbf{E2E2} & \textbf{3.22} & \textbf{2.74} & \textbf{2.79} & \textbf{2.80} & \textbf{2.76} & \textbf{2.79} & \textbf{17.09} \\
E2E3 & 3.20 & 2.73 & 2.91 & 2.76 & 2.85 & 2.83 & 17.28 \\
E2E4 & 3.22 & 2.85 & 2.89 & 2.87 & 2.87 & 2.81 & 17.50 \\
E2E5 & 3.24 & 2.81 & 2.85 & 2.86 & 2.87 & 2.88 & 17.52 \\
\hline
Mean & 3.22 & 2.79 & 2.85 & 2.82 & 2.84 & 2.83 & 17.35 \\
\hline
\hline
INDI1 & 3.30 & 2.86 & 2.80 & 2.86 & 2.91 & 2.79 & 17.51 \\
\textbf{INDI2} & \textbf{3.26} & \textbf{2.79} & \textbf{2.80} & \textbf{2.79} & \textbf{2.84} & \textbf{2.77} & \textbf{17.26} \\
INDI3 & 3.25 & 2.80 & 2.83 & 2.86 & 2.76 & 2.84 & 17.35 \\
INDI4 & 3.27 & 2.78 & 2.82 & 2.76 & 2.83 & 2.83 & 17.29 \\
INDI5 & 3.31 & 2.80 & 2.86 & 2.81 & 2.87 & 2.94 & 17.58 \\
\hline
Mean & 3.28 & 2.80 & 2.82 & 2.82 & 2.84 & 2.83 & 17.40 \\
\bottomrule
\end{tabular}

    \caption{Lap times (in seconds) fastest flight in bold.}
    \label{tab:lap_times}
\end{table}

\section{CONCLUSION}
We have presented a novel neural network approach for achieving high-speed quadcopter flight through end-to-end reinforcement learning. We addressed the reality gap issue by using an adaptive method alongside a learned residual model. Comparing our approach to a state-of-the-art INDI network, we observed that our E2E network can outperform the state of the art both in simulation in real flights. 
In the simulated environment, E2E exhibits a notable advantage, completing tasks 1.39 seconds faster. This performance gap, while smaller in real flight scenarios, remains significant and showcases the competitiveness of E2E architectures when compared to the state of the art.
The performance contrast observed between simulation and real-life scenarios implies that our E2E approach might exhibit greater sensitivity to modeling errors, as opposed to the robustness of the INDI controller. Future efforts could be directed toward enhancing the E2E network's ability to adapt to modeling errors. Additionally, apart from addressing moment and thrust offsets, it's also possible to identify other discrepancies in the model, such as variations in battery voltage or maximum RPM. Alternatively, a promising avenue for future research may involve refining our network through the application of offline RL techniques using real flight data.

\addtolength{\textheight}{-12cm}   










\bibliographystyle{IEEEtran}
\bibliography{root}

\end{document}